%% file: root.tex
\documentclass[letterpaper, 10 pt, conference]{ieeeconf}  % Comment this line out if you need a4paper

\IEEEoverridecommandlockouts                              % This command is only needed if 
\let\labelindent\relax
\usepackage{enumitem}
\usepackage{balance}
\usepackage[font={small}]{caption}
\usepackage[labelformat=simple]{subfig}
\usepackage{array}
\usepackage{textcomp}
\usepackage{mathtools, nccmath}
\usepackage{graphicx}
\usepackage{amsfonts}
\usepackage{amsmath}
\usepackage{amssymb}
\usepackage{hyperref}
\usepackage{tikz}
\usepackage{arydshln}
\usepackage{multirow}
\usepackage{bm}
\usepackage{epstopdf}
\usepackage{cite}
\usepackage{siunitx}
\usepackage{xcolor}
\usepackage[ruled,linesnumbered,boxed,noend]{algorithm2e}
\usepackage{booktabs}
\usepackage{graphicx}
\usepackage{threeparttable}
\usepackage{stfloats}
\usepackage{cases}

\title{\LARGE \bf
Dynamic Control Barrier Function-based Model Predictive Control to Safety-Critical Obstacle-Avoidance
of Mobile Robot
}

\author{Zhuozhu Jian$^\dagger$, Zihong Yan$^\dagger$, Xuanang Lei, Zihong Lu,\\Bin Lan*, Xueqian Wang*, Bin Liang% <-this % stops a space
\thanks{$^\dagger$ indicates equal contribution.}%<-this stops a blank space
\thanks{* Corresponding authors: Xueqian Wang,Bin Lan.}
\thanks{This work is supported by the National Natural Science Foundation of China (Grant No. U21B6002, U1813216)}
\thanks{Zhuozhu Jian, Zihong Yan, Bin Lan, Xueqian Wang, and Bin Liang are with the Center for Artificial Intelligence and Robotics, Shenzhen International Graduate School, Tsinghua University, Shenzhen 518055, China, \tt\{jzz21@mails., yanzh22@mails., lan.bin@sz., wang.xq@sz., liangbin@\}tsinghua.edu.cn}
\thanks{Xuanang Lei is with Department of D-MAVT at ETH Zurich, Zurich 8092, Switzerland. \tt xualei@student.ethz.ch}
\thanks{Zihong Lu is with School of Mechanical Engineering and Automation at Harbin Institute of Technology, Shenzhen 518055, China. \tt 200320802@stu.hit.edu.cn}
}

\begin{document}

\maketitle
\thispagestyle{empty}
\pagestyle{empty}

%%%%%%%%%%%%%%%%%%%%%%%%%%%%%%%%%%%%%%%%%%%%%%%%%%%%%%%%%%%%%%%%%%%%%%%%%%%%%%%%
\begin{abstract}
This paper presents an efficient and safe method to avoid static and dynamic obstacles based on LiDAR. First, point cloud is used to generate a real-time local grid map for obstacle detection. Then, obstacles are clustered by DBSCAN algorithm and enclosed with minimum bounding ellipses (MBEs). In addition, data association is conducted to match each MBE with the obstacle in the current frame. Considering MBE as an observation, Kalman filter (KF) is used to estimate and predict the motion state of the obstacle. In this way, the trajectory of each obstacle in the forward time domain can be parameterized as a set of ellipses. Due to the uncertainty of the MBE, the semi-major and semi-minor axes of the parameterized ellipse are extended to ensure safety. We extend the traditional Control Barrier Function (CBF) and propose Dynamic Control Barrier Function (D-CBF). We combine D-CBF with Model Predictive Control (MPC) to implement safety-critical dynamic obstacle avoidance. Experiments in simulated and real scenarios are conducted to verify the effectiveness of our algorithm. The source code is released for the reference of the community\footnote{Code: \url{https://github.com/jianzhuozhuTHU/MPC-D-CBF}.}.

% Autonomous navigation of ground robots has been widely used in static environments, but there are still many challenges in dynamic and crowded environments where obstacles are moving fast. The paper proposed a safety-critical Model Predictive Control (MPC) utilizing discrete-time Dynamic Control Barrier Functions (DCBFs). We cluster moving obstacles as elliptical cylinders and predict their shape-and-position changes in a few steps in the future, given by an adaptive Kalman Filter (KF) algorithm. Experiments in real scenarios are conducted to verify the effectiveness of the method. Our method is fully onboard and can be applied to other autonomous vehicles or flights.

\end{abstract}

%%%%%%%%%%%%%%%%%%%%%%%%%%%%%%%%%%%%%%%%%%%%%%%%%%%%%%%%%%%%%%%%%%%%%%%%%%%%%%%%

\input{sections/introduction.tex}

\input{sections/overview.tex}

\input{sections/implementation.tex}

\input{sections/experiments.tex}

\input{sections/conclusions.tex}

%%%%%%%%%%%%%%%%%%%%%%%%%%%%%%%%%%%%%%%%%%%%%%%%%%%%%%%%%%%%%%%%%%%%%%%%%%%%%%%%

\newpage
{
    \bibliographystyle{IEEEtran}
    \bibliography{IEEEabrv, bib/bibliography}
}

\end{document}

%% file: sections/introduction.tex
\section{INTRODUCTION}

Research on the safe-critical optimal planning and control of mobile robots has been actively conducted during the past decades. By improving the robustness and efficiency of existing autonomous navigation solutions \cite{putz2018move}\cite{bansal2020combining}\cite{kalogeiton2019real}, the autonomous navigation of robots has been widely used in many fields \cite{yuan2019multisensor}\cite{xiao2021robotic}, but autonomy in the dynamic and unstructured environment remains a challenge. The difficulties are mainly in the following aspects: 1) stable detection and prediction of obstacles in an unstructured environment; 2) parametric representation and uncertainty analysis of obstacles; 3) motion planning algorithm for real-time moving obstacle trajectories.

To address these issues, this paper presents an onboard lidar-based approach for safe navigation of vehicles in a dynamic environment. We detect obstacles based on real-time point cloud, perform the clustering operation, and then parameterize the obstacles as MBEs. To deal with the uncertainty of MBE, we summarize the influence of MBE shape parameters on MBE center position parameters, and use KF to realize the stable estimation of obstacles. Uncertainty is used to extend the obstacle to enhance safety. For motion planning in a dynamic environment, we propose D-CBF-based MPC to obtain a safe collision-free trajectory.

\begin{figure}[t]
    \centering
    \includegraphics[width=8.5cm]{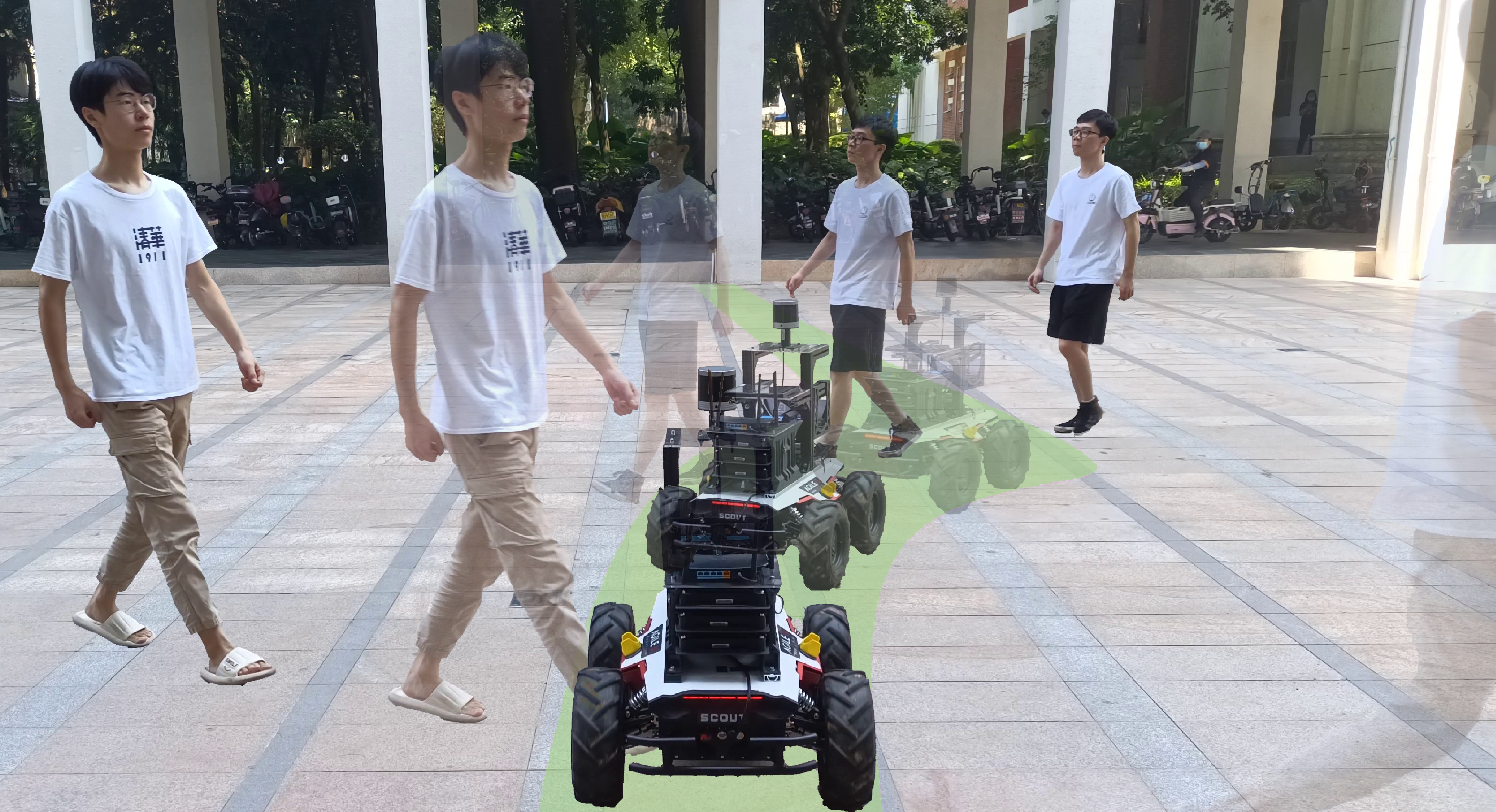}
    \caption{An example of a differential vehicle using our approach to avoid pedestrians. The Scout 2.0 four-wheel-drive platform is used to conduct experiments of avoiding more dynamic obstacles (such as electromobile and dogs).}
    \label{fig_exp_sen1}
    \vspace{-0.2cm}
\end{figure}

\subsection{Related Work}

\begin{figure*}[t]
    \center
    \includegraphics[width=16cm]{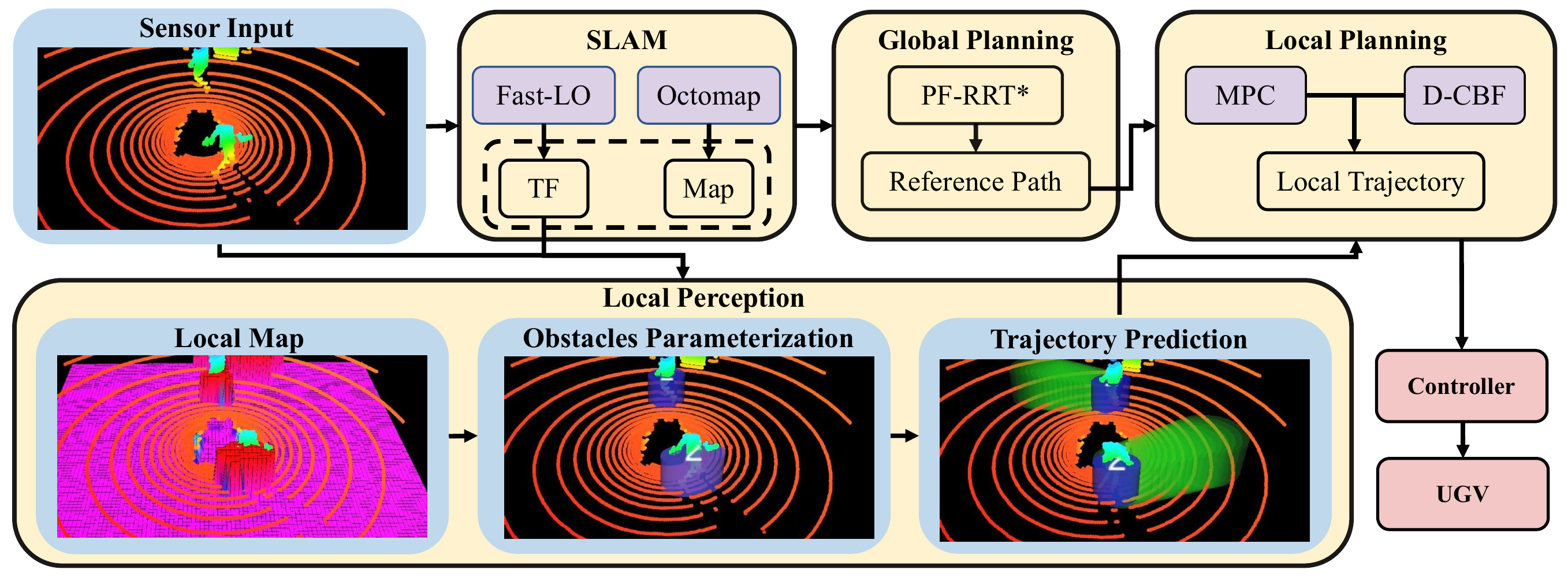}
    \caption{
    Overview of system framework. From left to right: LiDAR sensor detects the surrounding environment and generates point cloud. The SLAM module receives point cloud to generate TF and maintains a 3D probabilistic map in real time. The global planning module generates a reference path based on PF-RRT* and the map. Point cloud is used to generate the local map. Obstacles are parameterized on the local map by clustering to generate MBE. By using Kalman filter, the current motion state of MBE is estimated and then used to generate the corresponding trajectory in the forward time domain. The predicted trajectory and the reference path are sent to the local planning module for motion planning. This module combines D-CBF and MPC, generates the safe trajectory of the robot, and sends the movement command  in real time. The UGV receives motion commands to move along the trajectory. 
    }
    \label{fig_framework_1}
    \vspace{-0.3cm}
\end{figure*}

%  \textit{Obstacle Detection:} 
% To achieve critical safety navigation in dynamic environments, robots need to detect the obstacles around them efficiently and fast. One major method for detecting and tracking objects in dynamic scenarios is based on visual information from images obtained from cameras. \cite{rgbd}\cite{yolov3} However, these approaches cannot often detect fast-moving obstacles in real time. \cite{stereo} proposes a real-time algorithm to detect and track generic dynamic obstacles based on stereo camera data. The most visible deficiency is that it can only detect short distances and narrow scopes, which means the inability to avoid fast-moving objects a few meters away but about to collide in a short time. 
%  \textit{Trajectory Prediction:} 
% To avoid dynamic obstacles, predicting their positions in the future is also necessary. \cite{MPCmora}  estimates obstacle's future positions and uncertainties with a linear Kalman filter. 

 \textit{Local Perception:} 
To achieve safety-critical navigation in dynamic environments, robot needs to detect surrounding obstacles efficiently and fast. One method for detecting and tracking objects is based on visual information obtained from cameras \cite{rgbd}\cite{yolov3}\cite{stereo}\cite{lin2020robust}. However, This method does not perform well in bad lighting and weather conditions. Another method is based on point cloud data. 
% Current dynamic obstacle avoidance systems commonly leverage detection and tracking of obstacles with a Gaussian noise and modeled obstacles with certain shapes.
Wang et al. \cite{wang2021autonomous} developed a dynamic environment perception method and model dynamic obstacles with ellipsoids based on RGB-D camera, and proposed a efficient path searching method considering dynamic objects avoidance constraints.
Brito et al.\cite{MPCmora} detects and models dynamic obstacles as ellipses based on LiDAR sensor, and uses a linear Kalman filter to estimate their future positions with a constant velocity model with Gaussian noise in acceleration, and achieves stable dynamic obstacle avoidance. However, the uncertainty distribution of obstacles in the future time domain is assumed to be uniform in \cite{MPCmora}, which is improved in this paper

 \textit{Local Planning:} 
Collision avoidance in static and dynamic environments can be achieved through various methods, such as DWA (Dynamic Window Approach) \cite{fox1997dynamic}, artificial potential fields \cite{khatib1986real}, social forces \cite{ferrer2013robot}, gradient map \cite{gradient}, pre-computed motion primitives library \cite{aggressive3dcollision}\cite{funnel}, collision-free flight corridor\cite{MPCmora}\cite{gaofeicorridor}. However, these methods are not effective for higher speed obstacles or more complex environments. Our approach is based on MPC, which is more and more widely used in recent years \cite{hewing2020learning}\cite{berberich2020data}\cite{carli2020model}. Bruno Brito et al. \cite{MPCmora} proposed a planning approach based on Model Predictive Contouring Control (MPCC) in dynamic, unstructured environments. However, this algorithm only has good avoidance effect on pedestrians, and it is difficult to achieve similar effect on more complex obstacles (such as elastic balls). Zeng et al. \cite{zeng2021safety} proposed a model predictive control design and \cite{AnalysisMPCCBF} analysed its feasibility and safety, where discrete-time CBF constraints are used in a receding horizon fashion to ensure safety. However, this algorithm doesn't perform well in dynamic environment. Our approach extends CBF to achieve dynamic obstacle avoidance.

\subsection{Contribution}
The contributions of this paper are as follows.
\begin{itemize}
\item 
Dynamic Control Barrier Function-based Model Predictive Control framework is proposed to generate a safe collision-free trajectory in a dynamic environment.

\item 
% A method based on DBSCAN and Kalman filter is proposed to parameterize the obstacle as MBE and extend the safe boundary using uncertainty.
We propose a method for detection and prediction of obstacles based on minimum bounding ellipse and Kalman filter, in which uncertainty is considered to enhance safety.

\item Experiments in gazebo and real scenarios are carried out to verify the real-time performance, effectiveness and stability of the safe-critical obstacles avoidance algorithm.
\end{itemize}

%% file: sections/overview.tex
\section{OVERVIEW OF THE FRAMEWORK}

The algorithm flow of the system is shown in Fig.\ref{fig_framework_1}. SLAM module consists of Fast-LO \cite{zhu2022robust} and Octomap \cite{hornung2013octomap} , which are used to complete localization and mapping respectively. Local perception module detects obstacles and predicts their trajectories. First, a local map based on real-time point cloud data is generated, and then the obstacles are parameterized into MBE, which will be described in detail in \ref{subsec:Local map} and \ref{subsec:Obstacle parameterization}. We propose a KF-based method to predict the trajectory of the MBE and extend the MBE with uncertainty in \ref{subsec:Uncertainty analysis} and \ref{subsec:Trajectory prediction}. The local planning is implemented by MPC with D-CBF in \ref{subeq:dynamics} to generate the collision-free local trajectory. In this way, the robot completes safe and robust dynamic obstacle avoidance with only LiDAR as the sensor.

% The UGV is equipped with LiDAR to generate the point cloud. SLAM module consists of Fast-LO \cite{zhu2022robust} and Octomap \cite{hornung2013octomap} , which are used to complete localization and mapping respectively. The function of the local perception module is obstacle parameterization. First, a local map based on real-time point cloud data is generated, and then the obstacles are parameterized into elliptical cylinders and associated with the corresponding label. And this part will be described in detail in \ref{subsec:Local map} and \ref{subsec:Obstacle parameterization}. The Kalman filter is used to estimate the state of the elliptical cylinders and to predict the motion of obstacles in the forward time domain, which will be described in \ref{subsec:Trajectory prediction}. The local planning is implemented by Dynamic-MPC-CBF in \ref{subeq:dynamics}, which takes into account both the global reference trajectory and the obstacle trajectory to generate the Collision-free local trajectory. Thus, the robot can achieve safe and robust avoidance of dynamic obstacles in an entirely lidar-based perception.

%% file: sections/implementation.tex
\section{IMPLEMENTATION}

% \begin{figure*}[t]
%     \centering
%     \includegraphics[width=10cm]{figures/mpc-cbf.pdf}
%     \caption{Optimal path of robot $\mathbf{X}(t:t+N|t)$ to avoid collision with predicted obstacle $O_i(t:t+N|t)$ in $N$ future steps. By connecting the centers of the robot and ellipse, the distance between the periphery and the robot can be computed as $l_i(t:t+N|t)$. The set of distances sweeps the CBF zone and is constrained by CBF bounds, which prevents the robot from approaching the obstacle too fast.}
%     \label{fig:obs representation}
%     \vspace{-0.3cm}
% \end{figure*}

\subsection{Local Planning}
\subsubsection{Dynamic Control Barrier Function}
\
\newline
\indent
It has been proved that CBF(Control Barrier Functions) can be used for UGV obstacle avoidance\cite{zeng2021safety}\cite{glotfelter2019hybrid}\cite{singletary2021comparative}. However, for dynamic obstacles, the safety of traditional CBF remains a challenge. 
Unlike CBF, D-CBF (Dynamic Control Barrier Function) considers obstacles to be movable. Define robot position, obstacles position, obstacles shape as $\boldsymbol{x}\in \mathbb{R} ^{n}$, $\boldsymbol{x}_{ob}\in \mathbb{R} ^{m}$, $\boldsymbol{\eta}_{ob}\in \mathbb{R} ^{p}$, and variable $X=[\boldsymbol{x},\boldsymbol{x}_{ob},\boldsymbol{\eta}_{ob}]\in \mathbb{R} ^{n} \times \mathbb{R} ^{m} \times \mathbb{R} ^{p}$.

For safety-critical control, the set $C$ defined as the superlevel set of a continuously differentiable function $h:\mathcal{X} \subset \mathbb{R} ^{n} \times \mathbb{R} ^{m} \times \mathbb{R} ^{p} $:
\begin{equation}
	\begin{aligned}
		C=\left\{ X\in \mathcal{X} :h( X ) \ge 0 \right\} 
	\end{aligned}
\end{equation}
Throughout this paper, we refer to $C$ as the safe set. Referring to the definition of CBF, D-CBF can be defined: $\frac{\partial h}{\partial X}\ne 0$ for all $X\in \partial C$, and there exists an extend class $\mathcal{K} _{\infty}$ function $\gamma$
\begin{equation}
    \label{equ:D-CBF}
	\begin{aligned}
	\exists u\,\,s.t. \dot{h}\left( X,u \right) \ge -\gamma \left( h\left( X \right) \right) , \gamma \in \mathcal{K} _{\infty}
	\end{aligned}
\end{equation}
By definition, D-CBF is a CBF on $\mathcal{X}$. As a result of the main CBF theorem, the safe set $C$ is forward invariant and asymptotically stable \cite{ames2019control}. Therefore, the set of component corresponding to the robot position in $C$ is also invariant and asymptotically stable, which implies that the control system is safe. 

\begin{figure}[t]
    \centering
    \includegraphics[width=8.5cm]{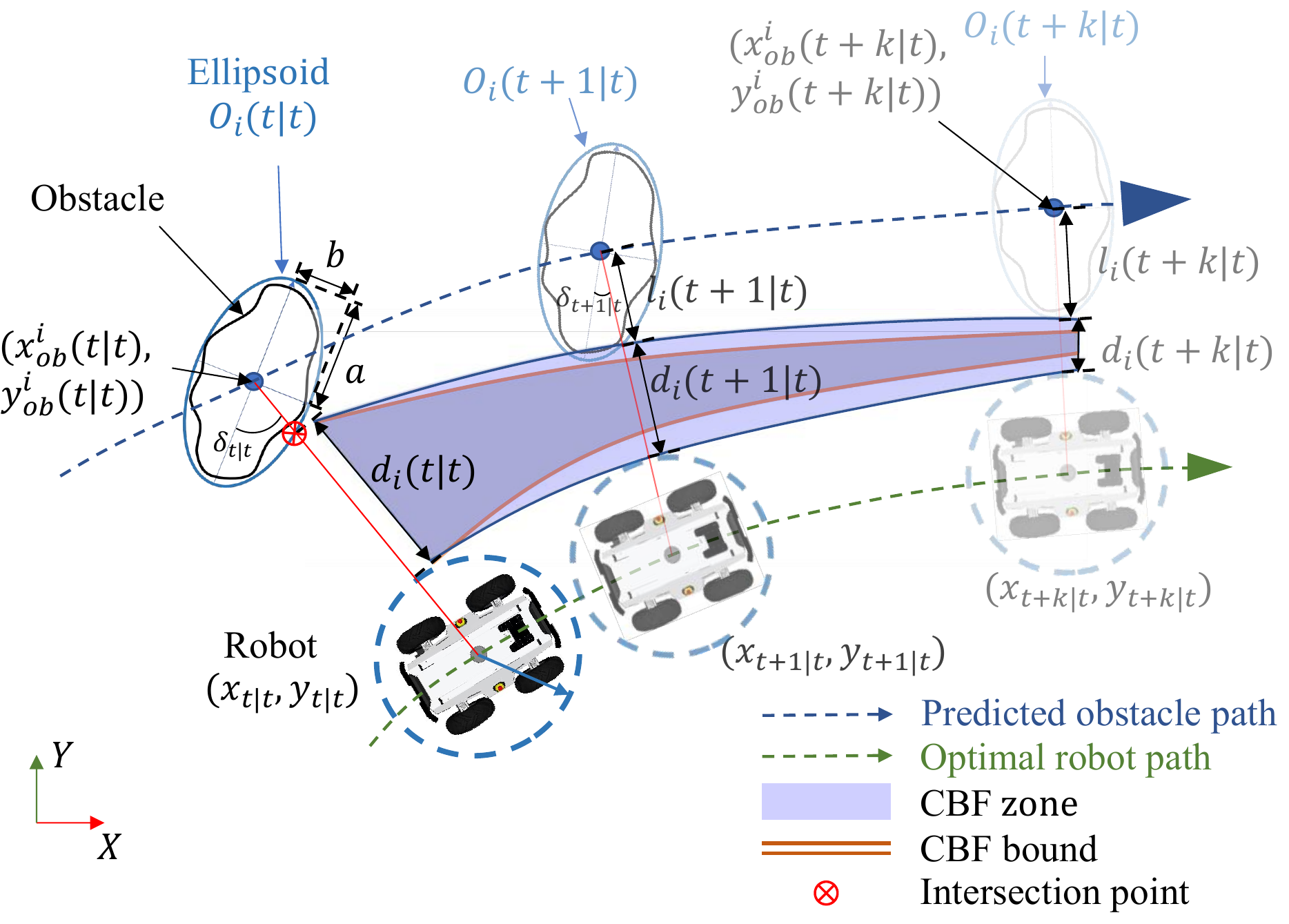}
    \caption{Optimal path of robot $\boldsymbol{x}(t:t+N|t)$ to avoid collision with predicted obstacle $O_i(t:t+N|t)$ in $N$ future steps. By connecting the centers of the robot and ellipse, the distance between the periphery and the robot can be computed as $d_i(t:t+N|t)$. The set of distances sweeps the CBF zone and is constrained by CBF bounds, which prevents the robot from approaching the obstacle too fast.}
    \label{fig:obs}
    \vspace{-0.4cm}
\end{figure}

In dynamic environment, D-CBF requires the input $\boldsymbol{u}$ to satisfy stricter constraints than CBF.
% In dynamic environment, additional constraints is required to avoid inequality \ref{equ:D-CBF} having no solution. 
For input-affine system $\dot{\boldsymbol{x}}=f\left( \boldsymbol{x} \right) +g\left( \boldsymbol{x} \right) \boldsymbol{u}$, inequality \ref{equ:D-CBF} is equivalent to
\begin{equation}
    \label{equ:D-CBF-2}
	\begin{aligned}
\mathop {sup} \limits_{\boldsymbol{u}\in \mathcal{U}}\left[ L_fh+L_gh\boldsymbol{u}+\underset{\epsilon}{\underbrace{\frac{\partial h}{\partial \boldsymbol{x}_{ob}}\frac{\partial \boldsymbol{x}_{ob}}{\partial t}}}+\underset{\varepsilon}{\underbrace{\frac{\partial h}{\partial \boldsymbol{\eta }_{ob}}\frac{\partial \boldsymbol{\eta }_{ob}}{\partial t}}} \right] +\gamma h\geqslant 0
	\end{aligned}
\end{equation}
where function $\gamma$ is a constant, $\epsilon$ and $\varepsilon$, respectively, represents the influence of the obstacle position and shape changes on the input. Define:
\begin{equation}
	\begin{aligned}
K_{cbf} =\left\{ \boldsymbol u\in \mathcal U:L_fh\left( \boldsymbol{x} \right) +L_gh\left( \boldsymbol{x} \right) \boldsymbol u+\gamma h\left( \boldsymbol{x} \right)  \ge 0 \right\} \\
K_{dcbf} =\left\{ \boldsymbol u\in \mathcal U:L_fh+L_gh\boldsymbol u+ \epsilon+\varepsilon+\gamma h \ge 0 \right\}
	\end{aligned}
\end{equation}
as control set of CBF and D-CBF, respectively. When $\epsilon+\varepsilon<0$, $K_{dcbf}\subseteq K_{cbf} $; when $\epsilon+\varepsilon>0$, $K_{cbf} \subseteq K_{dcbf} $; when $\epsilon+\varepsilon=0$, $K_{cbf}= K_{dcbf}$. 

% It can be seen that when $h$ satisfies the following two restrictions, can ensure the inequality has a solution: 1) the change rate $\dot \eta_{ob}$ of obstacle shape is bounded; 2) velocity $\dot x_{ob}$ of obstacle is bounded. 

% It is easy to prove the sufficiency of these conditions. Since $\dot x_{ob}$ and $\dot \eta_{ob}$ are bounded, $(a)$ and $(b)$ are bounded. There always exists $u$ such that \ref{equ:D-CBF-2} has a solution. 

% Thus, a D-CBF satisfying the above condition guarantees that the inequality has a solution \ref{equ:D-CBF-2}. 

For discrete-time systems, inequality \ref{equ:D-CBF-2}:
\begin{equation}
	\begin{aligned}
		\varDelta h\left( X_k,u_k \right) \ge -\gamma h\left( X_k \right) , 0<\gamma \le 1
	\end{aligned}
\end{equation}
where $\varDelta h\left( X_k,u_k \right) :=h\left( X_{k+1} \right) -h\left( X_k \right)$.

Each dynamic obstacle could be represented as an ellipse of area $O_i \subset \mathbb{R}^2$ according to \ref{subsec:Obstacle parameterization}, which could be described as $\boldsymbol{x}_{ob}^i=\lbrack x_{ob}^i, y_{ob}^i \rbrack \in \mathbb{R}^2$ and $\boldsymbol{\eta}_{ob}^i=[a_i,b_i,\theta_i]\in \mathbb{R}^3$ in Fig.\ref{fig:obs}. At time step t, the set of all obstacles is described as $\mathcal{A}_t^{obs} = \bigcup_{i=1,2,...,n}O_i(t)$. According to \ref{subsec:Trajectory prediction}, the state of the forward time domain $N$ of the obstacle $O_i$ can be predicted as $O_i(t:t+N|t)$. To simplify notation, we use $k$ instead of $t+k|t$. By connecting the centers of $O_i(k)$ and the robot 
% $\mathbf{x}(t:t+N|t) = [x,y](t:t+N|t)$
$\boldsymbol{p}(k)=[x(k),y(k)]$ at time $k$, we compute the distance $l_i(k)$ from the $O_i(k)$ to its periphery by solving simultaneous equations of line and ellipse:
%%
%\begin{numcases}{}
%\frac{(x-x_{ob}^i(k))^{2}}{a(k)^2}+\frac{(y-y_{ob}^i(k))^{2}}{b(k)^2}=1\\
%(y-y_{ob}^i(k)) = (x-x_{ob}^i(k))\cdot\tan{\theta}\\
%l_i(k) = \sqrt{(x-x_{ob}^i(k))^2+(y-y_{ob}^i(k))^2}
%\end{numcases}
%where $\theta$ represents the angle formed by the line and the major axis %of ellipse. $l_i(t:k)$ can be obtained:
%%
\begin{equation}
	\begin{aligned}
		 l_i(k) = \sqrt{\frac{a_i^2(k) b_i^2(k)(1+\tan^2\delta_i(k))}{b_i^2(k)+a_i(k)^2\tan^2\delta_i(k)}},  \\
           for \ k = 0,1,...,N-1
	\end{aligned}
\end{equation}
% Then the distance between the obstacle $O_i(k)$ and the robot $\boldsymbol{x}(k)$ is:
% \begin{equation}
% 	\begin{aligned}
% 		 d_i(k) = \left\| \boldsymbol{x}\left( k \right) -\boldsymbol{x}_{ob}^{i}\left( k \right) \right\| _2, \\
%           for \ k = 0,1,...,N-1
% 	\end{aligned}
% \end{equation}
% where $\Delta{x_k} = x_{t+k|t}-x_{ob}^i(t+k|t)$, $\Delta{y_k}=y_{t+k|t}-y_{ob}^i(t+k|t)$.
where $\delta$ represents the angle formed by the line and the major axis of ellipse. Thus, D-CBF is formulated in a quadratic form as follows:
\begin{equation}
	\begin{aligned}
		h(X_{k}) = \left\| \boldsymbol{p}\left( k \right) -\boldsymbol{x}_{ob}^{i}\left( k \right) \right\| _2-l_i(k)-d_{safe}, \\
           for \ k = 0,1,...,N-1
	\end{aligned}
\end{equation}
where $d_{safe}$ is safe distance.  
% $l_i(t:t+N|t)$  can be calculated as:
% \begin{equation}
% 	\begin{aligned}
% 		 l_i(t+k|t) = \sqrt{\frac{a_i^2(t+k|t) b_i^2(t+k|t)(1+\tan^2\theta_i(t+k|t))}{b_i^2(t+k|t)+a_i(t+k|t)^2\tan^2\theta_i(t+k|t)}},  \\
%           for \ k = 0,1,...,N-1
% 	\end{aligned}
% \end{equation}

\subsubsection{Model Predictive Control}
\

To formulate the MPC, we first describe the robot's dynamic model by a discrete-time equation $\boldsymbol{x}_{t+1} = f(\boldsymbol{x}_t,\boldsymbol{u}_t)$,  where $\boldsymbol{x}_t \in \mathcal{X} \subset \mathbb{R}^n$ is the state of the system at step $t \in \mathbb{Z}^+$, $\boldsymbol{u}_t \in \mathcal{U} \subset \mathbb{R}^m$ represents the control input, and $f$ is locally Lipschitz as the differential-drive system model:
\begin{equation}
     \boldsymbol{x}_{t+1}
        = \boldsymbol{x}_t + 
        \begin{bmatrix}
            \cos\vartheta_t & 0 \\
            \sin\vartheta_t & 0 \\
            0&1\\
        \end{bmatrix}
        \boldsymbol{u}_{t}\Delta{t}
\end{equation}
note that here $\boldsymbol{x}_t=[x_t,y_t,\vartheta_t]^T$, and $\vartheta_t$ represents the robot direction.

The MPC control problem is given by a receding horizon optimization:

\begin{subequations}
\label{eq:optimization-formulation}
    \begin{align}
        J_t^*(\boldsymbol{x}_t) = \min_{\boldsymbol{u}_{t:t+N-1|t}} 
         p\ &(\boldsymbol{x}_{t+N|t}) + \sum_{k=0}^{N-1}q(\boldsymbol{x}_{k},\boldsymbol{u}_{k}) 
            \label{subeq:opti} \\
        \text{s.t.} \
        \boldsymbol{x}_{k+1} = f(\ & \boldsymbol{x}_{k},\boldsymbol{u}_{k}), k = 0,...,N-1  
            \label{subeq:dynamics} \\
        \boldsymbol{x}_{k} \in  \mathcal{X},\ & \boldsymbol{u}_{k} \in \mathcal{U}, k = 0,...,N-1
            \label{subeq:scope} \\
        \ & \boldsymbol{x}_{t|t} = \boldsymbol{x}_t,
            \label{subeq:initial}\\
        \ & \boldsymbol{x}_{t+N|t} \in \mathcal{X}_f,
            \label{subeq:end}\\
        \Delta{h}(X_{k},\boldsymbol{u}_{k}) \geq\ & -\gamma h(X_{k}), k = 0,...,N-1
            \label{subeq:cbf}
    \end{align}
\end{subequations}
where \ref{subeq:dynamics} describes the system dynamics, and \ref{subeq:initial} represents the initial condition at step $t \in \mathbb{Z}^+$. $\mathcal{X}$, $\mathcal{U}$ and $\mathcal{X}_f$ are the set of admissible states, inputs and terminal states, respectively. To track the reference path $\boldsymbol{x}_{t:t+N}^d$ given by the global planner. In Eq.\ref{subeq:opti},  the terminal cost $p(\boldsymbol{x}_{t+N|t}) := \left\|\boldsymbol{x}_{t+N|t}-\boldsymbol{x}_{t+N|t}^d\right\|_P$, and the stage cost $  q(\boldsymbol{x}_{k},\boldsymbol{u}_{k}):=\left\| \boldsymbol{x}_{k}-\boldsymbol{x}_{k}^{d} \right\| _Q+\left\| \boldsymbol{u}_{k} \right\| _R+\left\| \boldsymbol{u}_{k+1}-\boldsymbol{u}_{k} \right\| _S $, where $P$,$Q$,$R$,$S$ are the corresponding weight matrices.

\subsection{Local Perception}
\label{sec:Local Perception}

The local perception maintains and estimates the safety space and obstacle regions and predicts trajectory of obstacle while considering the uncertainty of observation. The overview of the local perception is shown in Fig.\ref{fig:localmap2}.
\subsubsection{Local map}
\label{subsec:Local map}
\
\indent 

\begin{figure}[t]
    \centering
    \includegraphics[width=8.5cm]{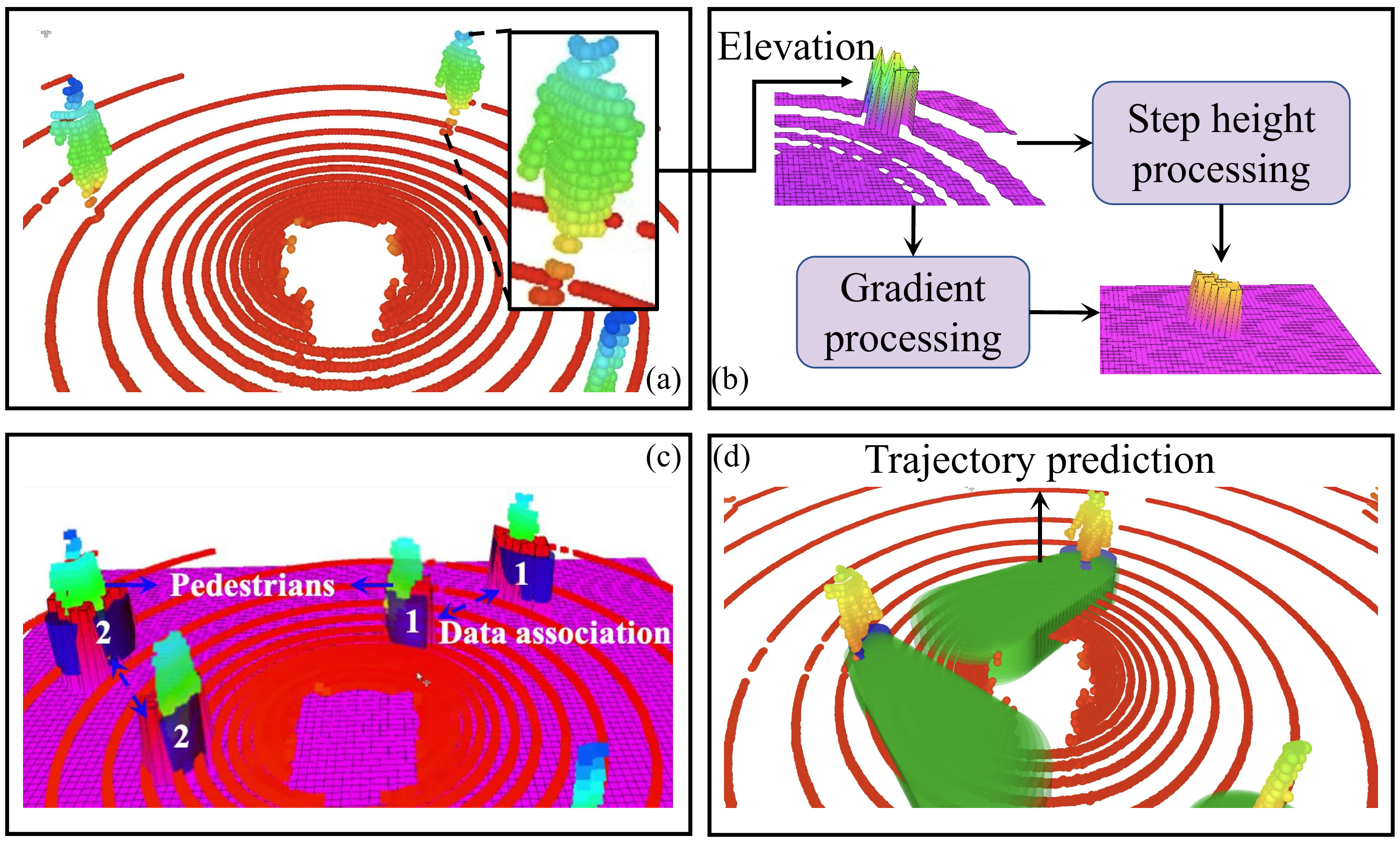}
    \caption{
    The overview of local perception. (a) Crop point cloud generated from LiDAR. (b) Local map representation and obstacle processing. (c) Obstacle clustering and enclosing with MBE and tracking between frames. (d) Trajectory prediction of obstacle involving uncertainty of observation. 
    }
   % \caption{Representation of trajectory for motion planning in complex terrain.}
    \label{fig:localmap2}
    \vspace{-0.2cm}
\end{figure}

To facilitate processing and maintain real-time performance, we use the point cloud generated from LiDAR to build a 2.5D elevation map based on gridmap\cite{fankhauser2016universal} representing the environment around the robot. 
Firstly, The local map is centered on the robot base and the point cloud is cropped via a passthrough filter to achieve real-time performance. Then the elevation of the cropped data is extracted and transformed to the local map. As the navigation of UGV pays more attention to the surface conditions on which they travel, we consider the maximum gradient G and step height $h_{step}$ of the robot operation for obstacle determination. The $3\times3$ Sobel operator $S_x$, $S_y$ and $3\times3$ Laplace operator $L$ are used for the convolution operation of the local map, suppose $(a,b)$ is a grid of the local map, and $M$ represents the elevation of its $3\times3$ neighborhood:

\begin{equation}
\begin{aligned}
G_S(a,b)= \sqrt{(S_x*M)^2+(S_y*M)^2}
\end{aligned}
\end{equation}

\begin{equation}
\begin{aligned}
G_L(a,b)= \left|L*M\right|
\end{aligned}
\end{equation}
while the step height $h_{step}(a,b)$ is directly obtained through adjacent cells of $(a,b)$ in the local map. For local obstacle representation, we threshold the local map as:

\begin{equation}
obs(a,b)=\left\{
\begin{array}{rcl}
1&& G_S(a,b)>S_{max} \vee G_L(a,b)>L_{max}\\ & & \vee h_{step}(a,b)>h_{max}\\
0&& {\text { others }}
\end{array} \right.
\end{equation}
where $S_{max}$, $L_{max}$ and $h_{max}$ represents the maximum operating limits of the robot, respectively.
As shown in Fig.\ref{fig:localmap2} (b), the obstacle regions are given a larger elevation value for visualization.

\subsubsection{Obstacle parameterization}
\label{subsec:Obstacle parameterization}
\

Considering obstacle avoidance in navigation tasks, it is meaningful to consider the minimum bounding volume of the obstacle. If the robot finds a feasible motion in the environment after the obstacle simplification, then this motion must also be feasible in the original space\cite{welzl1991smallest}. Therefore, the obstacles are clustered and enclosed with minimal bounding ellipses (MBE) in this module. In addition, we track the obstacle with a consistent label from historical data for trajectory prediction.
 
Firstly, Density-based clustering method DBSCAN \cite{ester1996density} is performed on the obstacle regions, resulting in a set of $n$ clusters $\mathcal{A}^{obs} = \{O_1, O_2, ..., O_n\}$.  And then the minimum bounding ellipse algorithm \cite{welzl1991smallest} is used to simplify each obstacle cluster $O_i$ to an ellipse $\xi$. In general, we parameterize the ellipse by its central coordinate, semi-major axis, semi-broken axis and rotation angle, i.e. $\xi=[x_{ob}, y_{ob}, a, b, \theta]$.
To associate the MBEs $E_{last}^m=\{\xi^1,\xi^2,...,\xi^m\}$ of last frame with the current frame $E_{cur}^n=\{\xi^1,\xi^2,...,\xi^n\}$, we first construct the affinity matrix $M_{m*n}$ by computing the distance $d$ between the center of $E_{last}^i$ and $E_{last}^j$. Then we use the Kuhn–Munkres algorithm \cite{kuhn1955hungarian} to complete the data association. Moreover, we reject a matching if the distance $d$ is larger than the max threshold $d_{max}$ and the unmatched ellipses in the new frame are assigned with a new label. As shown in Fig.\ref{fig:localmap2} (c), the pedestrians are tracked with consistent labels $1$ and $2$, respectively.

\subsubsection{Uncertainty analysis}
\label{subsec:Uncertainty analysis}
\

% In the process of robot trajectory planning, its trajectory and the corresponding uncertainty should be known for the moving obstacles. Based on KF (Kalman Filter), we propose a method to estimate the motion state of elliptic obstacles generated by clustering. 
% Different obstacles (vehicle, pedestrian, dog, etc.) have different movement intentions. However, in the limited range of the local map, the range of the future motion of the obstacle can still be represented by the uniform accelerated motion with corresponding uncertainty expansion. In \ref{subsec:Local Map}, obstacle is parameterized to MBE $\xi =\left[ x,y,a,b,\theta \right]\in \mathbb{R} ^5 $. 

Due to the noise of the point cloud and the change of detection angle of the obstacle, the shape $\boldsymbol{\eta}_{ob} =[a,b,\theta]$ of MBE can change rapidly and greatly, resulting in position $\boldsymbol{x}_{ob} =[x_{ob},y_{ob}]$ deviation from the real value.
%  \begin{figure}[htp]
%     \centering
%     \includegraphics[width=6.5cm]{figures/KF4.png}
%     \caption{
%     Change of MBE during a ball throwing movement.
%     }
%   % \caption{Representation of trajectory for motion planning in complex terrain.}
%     \label{fig:ball_throw}
%     \vspace{-0.2cm}
% \end{figure}
To address this issue, an indicator $\varXi_{p}$ for assessing MBE position confidence is proposed. 
\begin{equation}
\begin{aligned}
\varXi _{p}=\frac{\sum_{i=1}^m{\left\| \boldsymbol{x}_{ob,t-i,t}-\boldsymbol{x}_{ob,t-i,t}^{r} \right\| _{2}^{2}}}{m-1}
\end{aligned}
\end{equation}
where $\boldsymbol{x}_{ob,t-i,t}$ and $\boldsymbol{x}_{ob,t-i,t}^{r}$ represent MBE position and the real position of the obstacle at step $i$ in the backward time domain at time $t$, respectively. And $\varXi_{\eta}$ is proposed to assess the degree of shape change of the MBE.
% \begin{equation}
% \begin{aligned}
% \varXi _{\eta}=\frac{\sum_{i=1}^m{\left\| \eta_{t-i,t}-\eta_{t-i,t}^{r} \right\| _{2}^{2}}}{m-1}
% \end{aligned}
% \end{equation}
% where $\eta_{t-i,t}$ and $\eta_{t-i,t}^{r}$ represent the MBE shape and the real shape of the obstacle at step $i$ in the backward time domain at time $t$, respectively. 
In common scenarios, the shape of obstacles does not change rapidly (e.g., rapidly expanding). Therefore, in a limited area of the local map, the shape of the obstacle can be considered to remain unchanged. $\varXi_{\eta}$ can be estimated as follows:
\begin{equation}
\begin{aligned}
\varXi_{\eta}=\frac{\sum_{i=1}^m{\left\| \boldsymbol\eta_{ob,t-i,t}-\bar{\boldsymbol\eta}_{ob,t-m:t,t}
 \right\|_{2}^{2}}}{m-1}
\end{aligned}
\end{equation}
where $\bar{\boldsymbol\eta}_{ob,t-m:t,t}$ represents the mean of ${\boldsymbol\eta}$. 

\begin{figure}[t]
    \centering
    \includegraphics[width=8.5cm]{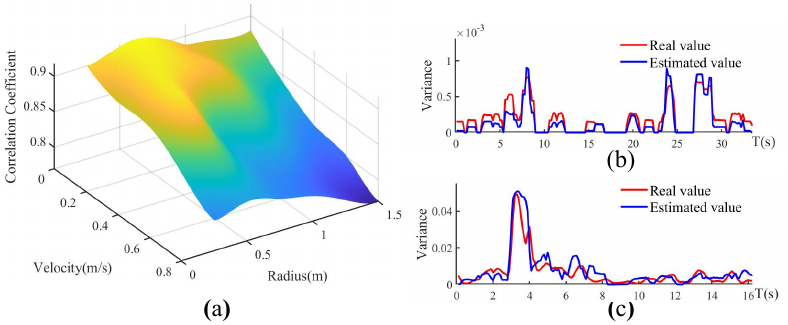}
    \caption{
    (a) shows the correlation coefficient of $\hat{\varXi}_{p}$ and ${\varXi}_{p}$ varies with obstacle speed and size when $\kappa$ is 5.5 and $\gamma$ is 1.3. (b)(c) show the changes of the true value ${\varXi}_{p}$ and estimated value $\hat{\varXi}_{p}$ of the confidence of cylindrical obstacles under stationary and moving conditions, respectively. In the experiment, the diameter of the cylinder was set as 1m, and the motion state is set as a uniform linear motion of 0.5m/s. The correlation sparsity of ${\varXi}_{p}$ and $\hat{\varXi}_{p}$ is 0.9133 when the obstacle is in a static state, and 0.9059 when the obstacle is in a state of constant velocity.
    }
   % \caption{Representation of trajectory for motion planning in complex terrain.}
    \label{fig:KF}
    \vspace{-0.4cm}
\end{figure}

Since $\boldsymbol x_{ob,t-i,t}^{r}$ is not available in a real environment, $\varXi _{\eta}$ can be used to estimate $\varXi _{p}$. Through data analysis, we get good estimation results using the formula $\hat{\varXi}_{p}=\kappa{\varXi}_{\eta}^{\gamma}$, in which 
both $\kappa$ and $\gamma$ are scale and power coefficients, respectively. To verify the correlation of $\hat{\varXi}_{p}$ and ${\varXi}_{p}$, experiments with obstacles of different speeds and sizes are conducted, as shown in Fig.\ref{fig:KF}. It can be seen that when $[\kappa,\gamma]=[5.5,1.3]$, this formula can provide a good estimate for obstacles with speed less than $1.5m/s$ and radius less than $0.9m$.

\subsubsection{Trajectory prediction}
\label{subsec:Trajectory prediction}
\ 
\newline
\indent 
In order to reduce the influence of point cloud noise, we use KF to update obstacles' states while using MBE $\xi$ as the observer. The state variable is set to a vector $\mathbf{x}
=\left[ \begin{matrix}{}
	x_{ob},y_{ob},a,b,\theta,	\dot{x}_{ob},		\dot{y}_{ob},		\ddot{x}_{ob},		\ddot{y}_{ob}\\
\end{matrix} \right] ^T \in \mathbb{R} ^9$, the state-transition matrix and the observation matrix are respectively: 

% \begin{equation}
% \begin{aligned}
% A=\left[ \begin{matrix}{}
% 	1&		0&		0&		0&		0&		T&		0&		\frac{T^2}{2}&		0\\
% 	0&		1&		0&		0&		0&		0&		T&		0&		\frac{T^2}{2}\\
% 	0&		0&		1&		0&		0&		0&		0&		0&		0\\
% 	0&		0&		0&		1&		0&		0&		0&		0&		0\\
% 	0&		0&		0&		0&		1&		0&		0&		0&		0\\
% 	0&		0&		0&		0&		0&		1&		0&		T&		0\\
% 	0&		0&		0&		0&		0&		0&		1&		0&		T\\
% 	0&		0&		0&		0&		0&		0&		0&		1&		0\\
% 	0&		0&		0&		0&		0&		0&		0&		0&		1\\
% \end{matrix} \right]\end{aligned}
% \end{equation}

% \begin{equation}
% \begin{aligned}
% H=\left[ \begin{matrix}{}
% 	1&		0&		0&		0&		0&		0&		0&		0&		0\\
% 	0&		1&		0&		0&		0&		0&		0&		0&		0\\
% 	0&		0&		1&		0&		0&		0&		0&		0&		0\\
% 	0&		0&		0&		1&		0&		0&		0&		0&		0\\
% 	0&		0&		0&		0&		1&		0&		0&		0&		0\\
% \end{matrix} \right]  \end{aligned}
% \end{equation}

\begin{equation}
\begin{aligned}
A_{9\times 9}=\left[ \begin{matrix}
	1&		0&		T&		0&		\frac{T^2}{2}&		0&		\\
	&		1&		0&		T&		0&		\frac{T^2}{2}&		\\
	&		&		1&		0&		T&		0&		O_{6\times 3}\\
	\vdots&		&		&		1&		0&		T&		\\
	&		&		&		&		1&		0&		\\
	0&		&		\cdots&		&		&		1&		\\
	&		&		O_{3\times 6}&		&		&		&		I_{3\times 3}\\
\end{matrix} \right] 
\end{aligned}
\end{equation}

\begin{equation}
\begin{aligned}
H_{5\times 9}=\left[ \begin{matrix}
	I_{2\times 2}&		O_{2\times 7}\\
	O_{3\times 6}&		I_{3\times 3}\\
\end{matrix} \right] 
\end{aligned}
\end{equation}

There are two reasons for the change of MBE position: (a) The movement of the obstacle itself. (b) Position deviation due to abnormal changes in MBE shape $\boldsymbol\eta_{ob}$. Based on the analysis in \ref{subsec:Uncertainty analysis}, a modified parameter $k$  for position covariance $R_{p}$ is proposed to reduce the deviation.

% In the second case, it is not accurate to use the position of the MBE to represent the real position of the obstacle. A modified parameter $k$ is proposed to update the covariance of the KF
\begin{equation}
\begin{aligned}
R_{p}=R_{p,\max}^{k}R_{p,\min}^{1-k}
\end{aligned}
\end{equation}
\begin{equation}
\begin{aligned}
k=\frac{\lg \left( \varXi _{p}/\varXi _{\min ,crit} \right)}{\lg \left( \varXi _{\max ,crit}/\varXi _{\min ,crit} \right)}\,\,  \,\end{aligned}
\end{equation}
where $R_{p,\max}$ and $R_{p,\min}$ represent the boundary of the position variance, $\varXi _{\min ,crit}$ and $\varXi _{\max ,crit}$ represent the boundary of $\varXi _{p}$. 
In this way, when the shape of MBE changes rapidly, $R_{p}$ will increase to reduce the vibration of MBE position.

Based on the current estimated state and the corresponding variance value, the state in future time domain can be predicted by $\mathbf{x}_k=A\mathbf{x}_{k-1}$ and $P_k=AP_{k-1}A^T+Q$, where $Q$ is the covariance of system noise.
% \begin{subequations}
% \label{eq:pre}
%     \begin{align}
    
%         \label{subeq:pre1} \\
    
%         \label{subeq:pre2} 
%     \end{align}
% \end{subequations}

In order to enhance safety, we use uncertainty to extend the ellipse. Ellipse $E_1=\mathrm{Diag}\left( \frac{1}{a^2},\frac{1}{b^2} \right) $ is enlarged by $\delta $ in both axis to $E_2=\mathrm{Diag}\left( \frac{1}{\left( a+\sigma \right) ^2},\frac{1}{\left( b+\sigma \right) ^2} \right) $. Define $r =r _{p}+r _{\eta}$ to describe the uncertainty of the ellipse, where $r _{p}$ and $r _{\eta}$ are obtained from the covariance of the position and shape of the ellipse, respectively. To find the smallest ellipse that bounds the Minkowski sum, the minimum value of $\sigma$ is obtained by solving the following equation \cite{kuhn1955hungarian}:
% To find the smallest ellipse that bounds the Minkowski sum, the minimum value of $\delta_{min}$ should be found. $\delta_{min}$ and $r$ have the following relationship \cite{kuhn1955hungarian}:
\begin{equation}
\begin{aligned}
\frac{2\left( \sigma _{\min}+r \right) ^2\left( \left( a+b \right) \left( \sigma _{\min}+r \right) +2ab \right)}{\left( a+b \right) \left( a+b+2\sigma _{\min}+2r \right)}-r^2=0
\end{aligned}
\end{equation}
For obstacles $O(t:t+N|t)$ in the predicted time domain, as $k$ increases from $0$ to $N$, the uncertainty of the obstacles increases, and the corresponding ellipse is expands accordingly, as shown in Fig.\ref{fig:localmap2}(d).

%% file: sections/experiments.tex
\begin{figure}[h]
    \centering
    \includegraphics[width=8.5cm]{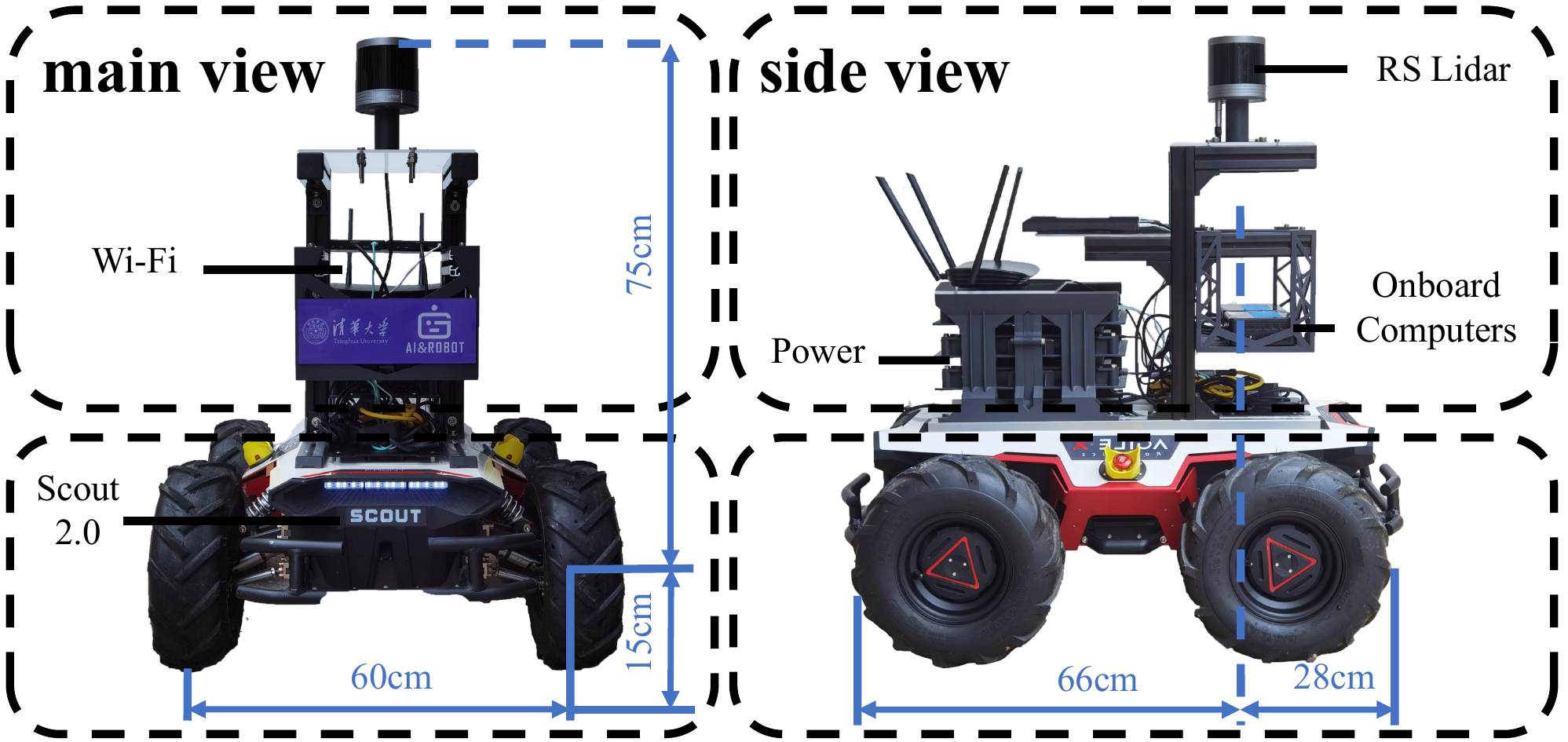}
    \caption{Robot platform for the experiment. Scout2.0, a four-wheel-differential-drive mobile robot equipped with battery pack, Wi-Fi is used for the real-world experiment. The only sensor used is RS-Helios, a 32-beam LiDAR, which has a vertical FoV of 70°, and 55°of FoV below horizon to eliminate blind zone. Two Intel@NUC with an i5 2.4GHz CPU and 16GB memory are used to run the planning algorithm and the SLAM algorithm, respectively.
    }
    \label{fig:robot}
    \vspace{-0.4cm}
\end{figure}

\section{EXPERIMENTS}
In this section, experiments in real scenarios and simulation environment are conducted to verify the effectiveness of our work. Our algorithm works under ROS Melodic operating system. The local perception module runs at 10-20 Hz, and the local planning module at 10 Hz.

\begin{figure*}[t]
    \center
    \includegraphics[width=17.8cm]{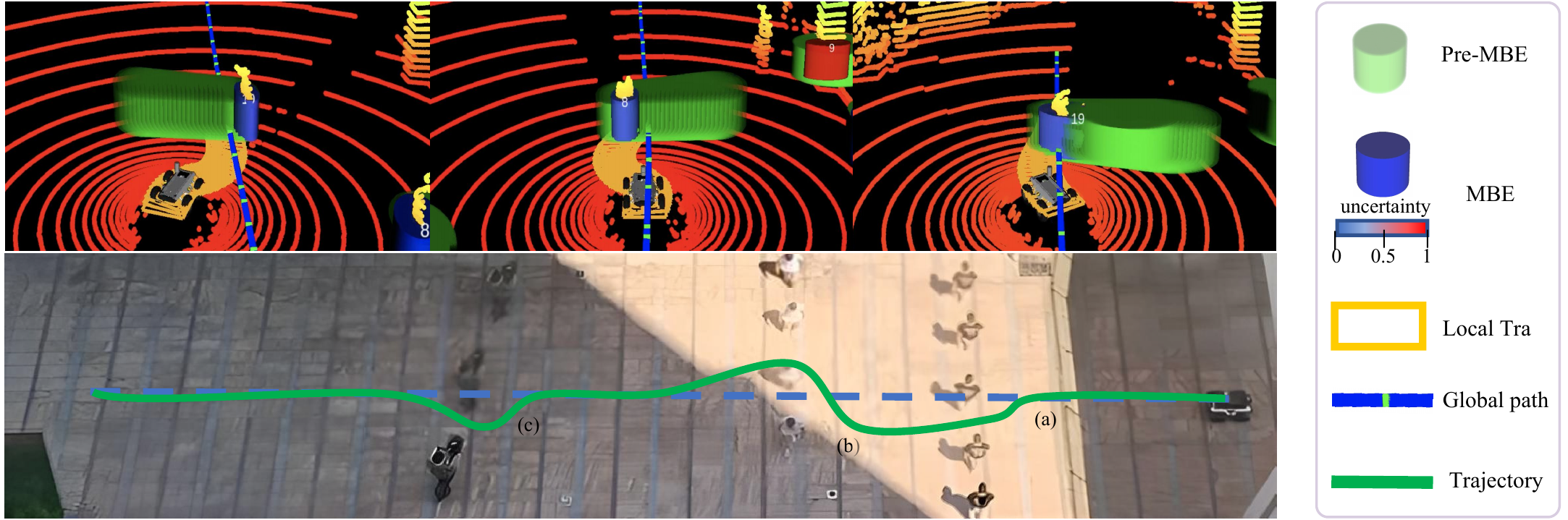}
    \caption{
    Real-world dynamic obstacle avoidance scenario. The blue line represents the global path generated by PF-RRT*\cite{jian2022putn}. The orange box represents the local trajectory executed by our MPC-D-CBF and the pose of robot footprint, respectively. The trajectories of the two pedestrians and the electromobile are demonstrated in the scenario, which are perceived as MBE with uncertainty from blue to red. The green cylinders represent the prediction of the obstacle trajectories. Figures (a),(b), and (c) on the top demonstrate the strategy of dynamic obstacles avoidance of our method in the real scene.
    }
    \label{fig:main-exp}
    \vspace{-0.5cm}
\end{figure*}

\subsection{Real-world scenarios}
% As shown in Fig.\ref{fig:robot}, Scout2.0, a four-wheel-differential-drive mobile robot is used for the real-world experiment. 
% % The only sensor used is RS-LiDAR-32, and two Intel NUC with an i5 2.4GHz CPU and 16GB memory is used for the SLAM perception and the robot motion planning, respectively.
% The only sensor used is RS-Helios, a 32-beam LiDAR, which has a vertical FoV of 70°, and 55°of FoV below horizon to elimilate blind zone. 

As shown in Fig.\ref{fig:main-exp}, a real-world experiment is conducted with two pedestrians and an electromobile. The distance from the starting point to the target point is set to $20m$, and the velocity of the pedestrian and the electromobile is about $1.2m/s$ and $1.8m/s$, respectively. The local perception range is set to $10 \times 10m$ with a resolution of 0.1m.
% , the neighborhood radius $eps$ and minimum number of neighbors $minpts$ of DBSCAN are set to $0.5m$ and $5$. 
The predicted step $N$ of the robot and the obstacle is set to $25$. Considering the blind spot of LiDAR, the safe distance $d_{safe}$ is $1.3m$. $\gamma$ in CBF is set to be $0.15$. 

% The robot navigate $20m$ from the left to right of the scenario with a global path generated by PF-RRT*, while avoiding collision with the dynamic objects with a velocity between $[0,1.2]$ $m/s$. The local perception range is set to $10 \times 10m$ to perceive and predict the obstacle motion. As the trajectory prediction of the obstacle shown in the Fig.\ref{fig:main-exp}, the velocities of the pedestrians and electromobile are about $1.5m/s$ and $2m/s$, respectively. 

In this scenario, when the robot encounters an obstacle (pedestrian,  electromobile), our algorithm generates a trajectory considering the predicted trajectory of the obstacle. For example, in Fig.\ref{fig:main-exp}(a), when the pedestrian is still on the left side of the robot, our algorithm predicts that its states in the next N steps will be a trajectory to the right. So the robot adopts the strategy of avoiding from the left. In this way, the robot can not only improve the safety of the movement, but also can reach the target point in a shorter time.
% When encountering dynamic obstacles, the robot navigates to the goal from the opposite direction of the obstacle trajectories according to their predicted velocities. If the speeds of the obstacles are lower to some extent, the robot will choose to pass in advance considering their velocities and positions.
In order to verify the scalability and robustness of our algorithm, experiments are carried out with the obstacles being elastic balls and a quadruped robot, which can refer to the attached video.

\subsection{Simulation scenario}

To evaluate the performance of the proposed method, we compare the proposed method with 4 baseline approaches: 
\begin{itemize}
\item 
MPC \cite{turri2013linear}: MPC tracking controller considering safety constraints with euclidean norm distance.

\item 
MPC-CBF\cite{zeng2021safety}: MPC tracking controller considering safety constraints with discrete-time CBF.

\item 
MPC-KF: MPC involving our KF obstacle trajectory prediction with uncertainty.

\item 
MPC-CBF-curvefit: MPC involving prediction of obstacle trajectory using polynomial curve fitting. 
\end{itemize}

We adopted the following indicators to compare the five algorithms: Min dist (minimum distance to the obstacle), Cons time (time spent from the start point to the end point), Reac time (time interval between the obstacle entering the local map and the robot's reaction), Speed var (speed variance when avoiding obstacles). The results are shown in Table \uppercase\expandafter{\romannumeral1}.

\begin{figure}
    \centering
    \includegraphics[width=8cm]{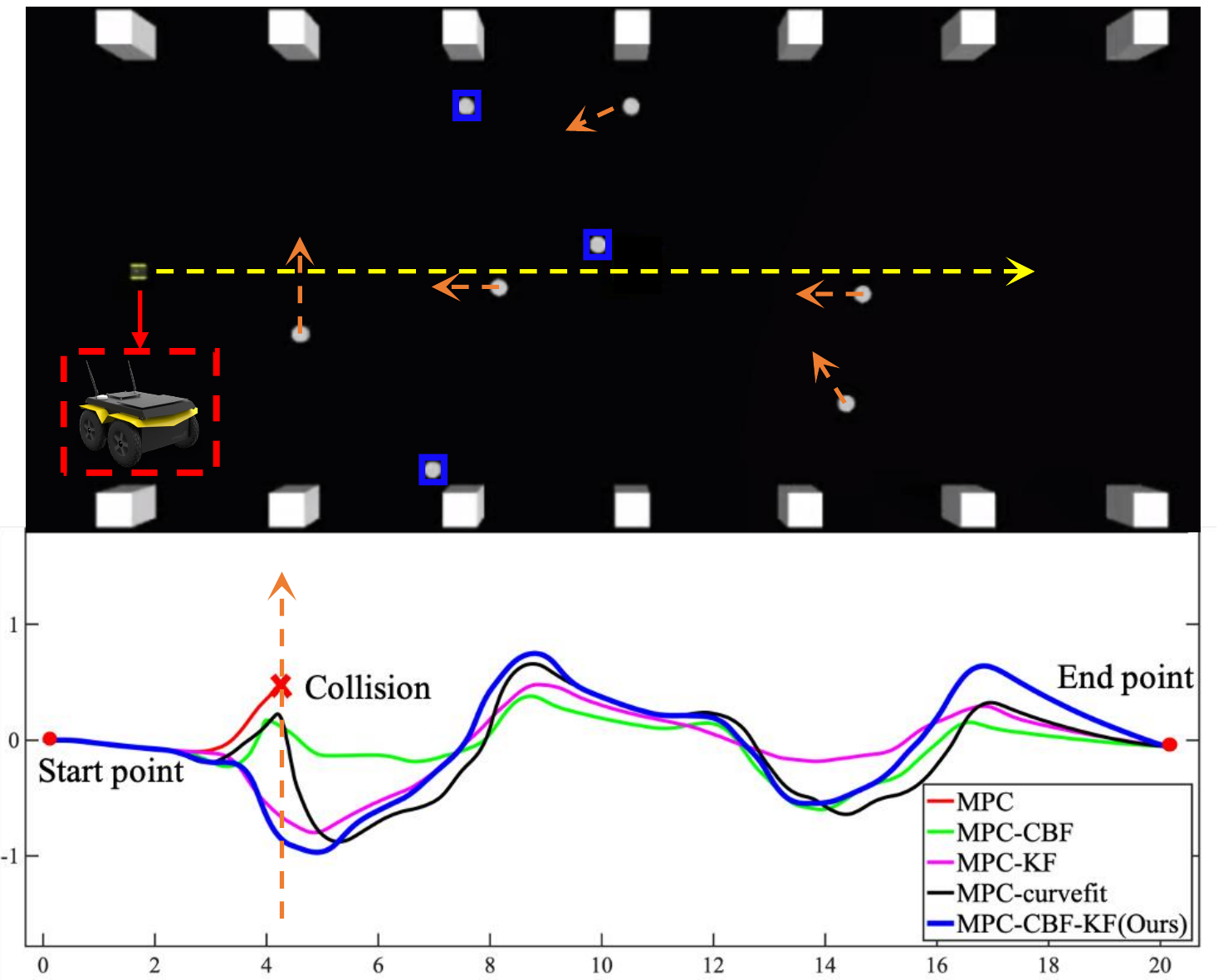}
    \caption{Simulation setup. An environment is built in Gazebo with dynamic obstacles. The autonomous ground robot Jackal equipped with a LiDAR Velodyne is used as the simulated robot. The robot follows a global path (yellow line) and avoids static obstacles (blue box) and dynamic obstacles (orange trajectory). The bottom figure represents the trajectory of the five approaches.}
    \label{fig:simulation}
    \vspace{-0.5cm}
\end{figure}

The trajectories generated by each algorithm are shown in Fig.\ref{fig:simulation} in different colors. The MPC approach (red) fails and collides with the first dynamic obstacle. MPC-CBF (green) and MPC-CBF-curvefit (black) cannot avoid the dynamic obstacles in advance due to no prediction or inaccurate prediction of obstacles caused by noises of sensor input or changes in observation. Our KF prediction algorithm considers the uncertainty of LiDAR data, for which the MPC-KF method (magenta) and our method (blue) can avoid dynamic obstacles smoothly and robustly in advance, exceeding other methods in React time and Speed var metrics as shown in Table \uppercase\expandafter{\romannumeral1}. In addition, compared with MPC-KF, our method balances safety of dynamic obstacle avoidance with speed, i.e. navigates safely and smoothly to the target position while maintaining a safer distance of Min dist as shown in Table \uppercase\expandafter{\romannumeral1}.
% In Table \uppercase\expandafter{\romannumeral1}, we benchmark these five algorithms in terms of minimal distance to the obstacle, time spent, react time, and speed variance near the obstacle. Speed variance reflects the smoothness of robot's velocity when bypassing obstacles. We observe that MPC-CBF-KF spends less time reacting to obstacles, is less likely to collide, and keeps the furthest distance to obstacles during the whole navigation process, proving itself more safe-critical than other methods.

\begin{table}[htb]   
\begin{center}   
\caption{Comparison with baseline methods.}  
\label{table:1} 
\tabcolsep=3mm
\begin{tabular}{m{2cm}<{\centering} m{1.3cm}<{\centering} m{0.8cm}<{\centering} m{0.7cm}<{\centering} m{0.7cm}<{\centering}}   
\hline   \textbf{Algorithms} & \textbf{Min dist(m)} & \textbf{Cons time(s)} & \textbf{ Reac time(s)} & \textbf{Speed var} \\   
\hline   MPC         & 0(collided) & - & 1.331 & -\\ 
\hline   MPC-CBF    & 0.273       & 22.07 & 0.357 & 0.1598\\  
\hline   MPC-KF      & 0.201       & 20.05 & 1.197 & 0.0061\\  
\hline   MPC-CBF-curvefit& 0.074       & 23.27 & 0.421 & 0.1306\\ 
\hline   Ours & 0.828       & 21.52 & 0.353 & 0.0172\\
\hline   
\end{tabular}   
\end{center}   
\end{table}

%% file: sections/conclusions.tex
\section{CONCLUSIONS}
This paper proposes a safety-critical method for static and dynamic obstacles avoidance based on LiDAR sensor. Firstly, we adopt a robust local perception module, which perceives and parameterizes obstacles to MBEs, and predicts obstacles trajectory considering uncertainty based on Kalman filter. Then, we combine Dynamic Control Barrier Function and Model Predictive Control framework to generate a safe collision-free trajectory for robot navigation in dynamic environment.
Real-world experiments avoiding different types of obstacles validate that our method is robust, safe, and allows running all algorithms on-board.  
Moreover, we compare our method with four baseline approaches (MPC, MPC-CBF, MPC-KF, and MPC-CBF-curvefit) in simulation. The experimental results show that our method is more safe and efficient than other methods in the process of obstacle avoidance.

% The experimental results demonstrate that our method spends less time reacting to obstacles, is less likely to collide, and keeps the furthest distance to obstacles during the whole navigation process, proving itself more safe-critical than the baseline methods. 